\documentclass[10pt,twocolumn,letterpaper]{article}

\usepackage{iccv}
\usepackage{times}
\usepackage{epsfig}
\usepackage{graphicx}
\usepackage{amsmath}
\usepackage{mathbbol}
\usepackage{amssymb}
\usepackage{booktabs,makecell}
\usepackage{verbatim}
\usepackage{multirow}
\usepackage{authblk}
\usepackage{blindtext}

\usepackage[breaklinks=true,bookmarks=false]{hyperref}

\iccvfinalcopy 

\def\iccvPaperID{7316} 

\ificcvfinal\fi

\begin{document}

\title{PG-RCNN: Semantic Surface Point Generation for 3D Object Detection}

\author{Inyong Koo\thanks{Denote equal contribution}
\quad Inyoung Lee\protect\CoauthorMark 
\quad Se-Ho Kim 
\quad Hee-Seon Kim 
\quad Woo-jin Jeon 
\quad Changick Kim}
\affil{
KAIST \\
Daejeon, South Korea\protect \\
\tt\small \{iykoo010, inzero24, ksh1040, hskim98, woojin.jeon337, changick\}@kaist.ac.kr
}
\newcommand\CoauthorMark{\footnotemark[\arabic{footnote}]}

\maketitle
 \ificcvfinal\thispagestyle{empty}\fi

\begin{abstract}
One of the main challenges in LiDAR-based 3D object detection is that the sensors often fail to capture the complete spatial information about the objects due to long distance and occlusion.
Two-stage detectors with point cloud completion approaches tackle this problem by adding more points to the regions of interest (RoIs) with a pre-trained network.
However, these methods generate dense point clouds of objects for all region proposals, assuming that objects always exist in the RoIs. This leads to the indiscriminate point generation for incorrect proposals as well.
Motivated by this, we propose Point Generation R-CNN (PG-RCNN), a novel end-to-end detector that generates semantic surface points of foreground objects for accurate detection.
Our method uses a jointly trained RoI point generation module to process the contextual information of RoIs and estimate the complete shape and displacement of foreground objects.
For every generated point, PG-RCNN assigns a semantic feature that indicates the estimated foreground probability.
Extensive experiments show that the point clouds generated by our method provide geometrically and semantically rich information for refining false positive and misaligned proposals.
PG-RCNN achieves competitive performance on the KITTI benchmark, with significantly fewer parameters than state-of-the-art models.
The code is available at \url{https://github.com/quotation2520/PG-RCNN}.

\end{abstract}


\section{Introduction}

3D object detection using LiDAR point clouds is a fundamental perception task in autonomous driving that has received significant attention in recent years.
LiDAR sensors are frequently used in many 3D applications, such as odometry and mapping \cite{loam_zhang, floam_wang, legoloam_shan}, object tracking \cite{pointtracknet_wang, tpn_wu, 3dmot_wu}, and detection \cite{second_yan, pvrcnn_shi, voxelrcnn_deng}, due to their ability to provide accurate distance information in various conditions.

LiDAR-based 3D object detectors use either a point-based \cite{pointnet_qi, pointnet++_qi, pointgnn_shi, pointformer_pan} or a voxel-based \cite{voxelnet_zhou, second_yan, votr_mao} network to generate bounding boxes for foreground objects. 
The two-stage framework with a proposal refinement stage is often adopted in many detectors to enhance the detection accuracy \cite{pointrcnn_shi, graphrcnn_yang}.
While researchers have explored different methods \cite{pvrcnn_shi, voxelrcnn_deng} to extract effective refinement features for the regions of interest (RoIs), some of the most recent works with voxel-based backbones \cite{ct3d_sheng, pdv_hu} revisit the point information within the RoIs at the refinement stage, considering the precise coordinates and density of internal points.

\begin{figure}[t]
\begin{center}
   \includegraphics[width=\linewidth]{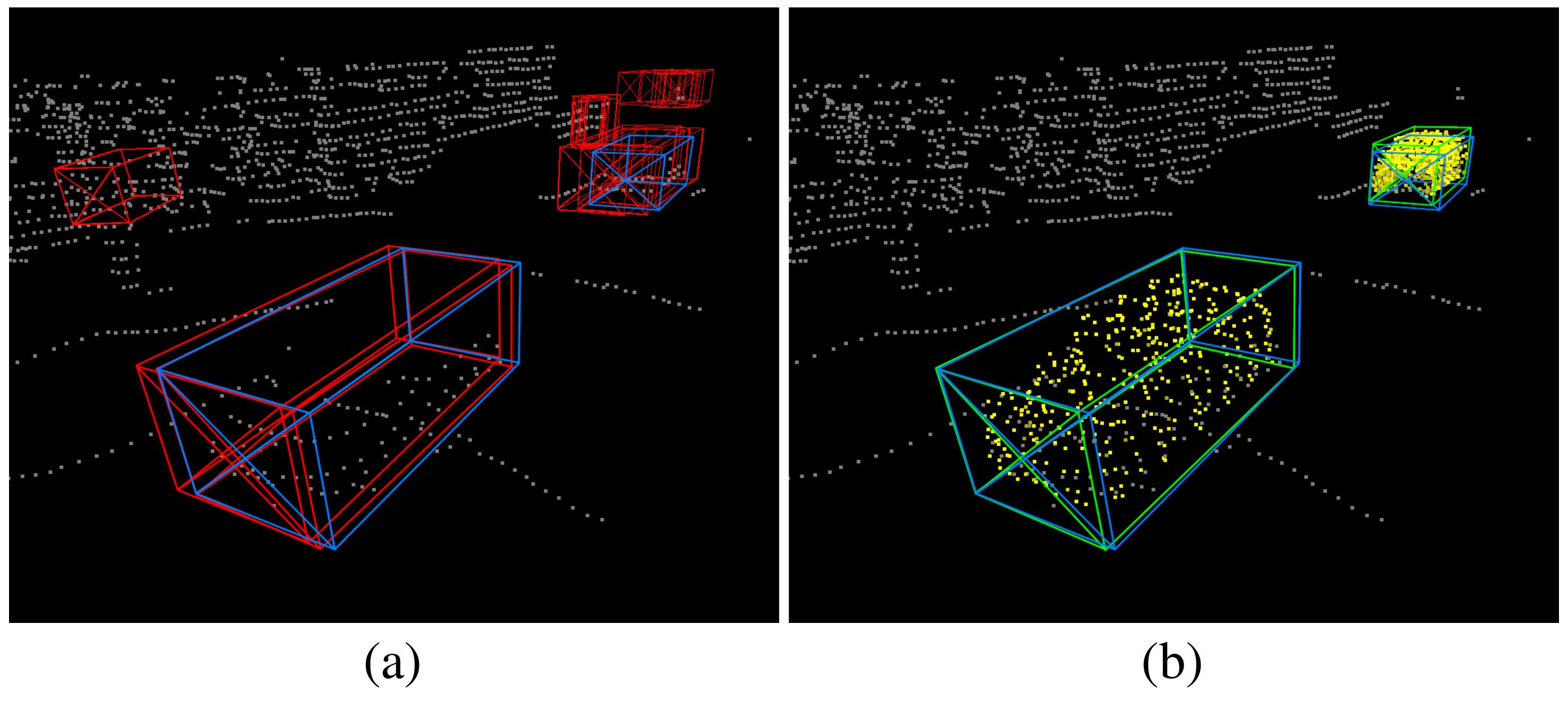}
\end{center}
    \vspace{-10pt}
   \caption{
   The intermediate outputs of PG-RCNN:  (a) first stage output with initial bounding box proposals in red, and (b) second stage outputs with generated points over 0.6 foreground score in yellow and final detection outputs in green bounding boxes. Ground truth bounding boxes are shown in blue.
   }
\label{fig:1_overview}
\end{figure}

Nevertheless, the inherent sparsity of LiDAR point clouds poses a challenge in 3D object detection, particularly for distant and occluded objects.
These objects have fewer collected points, making them difficult to detect and degrading the overall performance of detectors \cite{sienet_li, btc_xu}.
To address this problem, point cloud completion methods have been explored to assist proposal refinement by adding more points to the RoIs.
The methods in \cite{sienet_li, pcrgnn_zhang} enhance the resolution of point clouds by utilizing a point cloud completion network pre-trained from an external dataset \cite{shapenet}, but they only take point coordinates pooled from the proposal bounding box into account during the generation process.
As a result, they fail to capture the contextual information of the surroundings and indiscriminately produce dense point clouds for all proposals, including incorrect proposals.

Motivated by this, we propose the Point Generation R-CNN (PG-RCNN), an end-to-end two-stage 3D object detection method that can extract geometrically and semantically rich proposal refinement features via semantic surface point generation.
Our method includes the RoI point generation (RPG) module that estimates the actual shape and displacement of foreground objects, using primitive RoI features aggregated from the backbone as input.
Note that we jointly train the RPG module using auxiliary supervision from given data without introducing any external dataset.
While previous point cloud completion methods only output sets of spatial coordinates, our method goes beyond that by assigning a semantic feature to each generated point, which represents its estimated probability of belonging to the foreground.
These characteristics allow our novel point generation method to produce more informative point clouds for object detection.
Figure \ref{fig:1_overview} shows the intermediate outputs of our method.
PG-RCNN generates points with different foreground scores, presenting high-confidence foreground points for true positive proposals.
The generation points intuitively express the predicted location and shape of the objects, visualizing the reasoning process of PG-RCNN.

We demonstrate the effectiveness of our method with extensive experiments on the KITTI dataset \cite{kitti}. 
PG-RCNN achieves competitive performance with state-of-the-art models while significantly reducing the computational cost.
Qualitative results show that our approach better serves the purpose of refining false-positive or misaligned proposals compared to previous point cloud completion methods.

In summary, our main contributions are:
\begin{itemize}
    \item We present PG-RCNN, a novel two-stage 3D object detection method for LiDAR point clouds.
    In the proposal refinement stage, our method generates semantic surface points with foreground probabilities to extract shape-aware, semantically rich refinement features.
    \item We compare the point generation results of PG-RCNN to a previous point cloud completion approach and show that our method generates more effective points for object detection.
    \item PG-RCNN achieves competitive performance on the KITTI benchmark, with a significantly smaller number of parameters and inference time than the state-of-the-art models. 
\end{itemize}

\section{Related works}

\subsection{LiDAR-Based 3D Object Detection}

LiDAR-based 3D object detection methods can be categorized into two streams: point-based and voxel-based.
Point-based methods directly learn point features for detection by sampling raw point clouds and employing permutation-invariant operations.
The majority of point-based methods \cite{3dssd_yang, pointrcnn_shi, lidarrcnn_li, std_yang} use PointNet-like backbones \cite{pointnet_qi, pointnet++_qi}, while methods like \cite{pointgnn_shi, pointformer_pan} adopt other architectures to process the sampled points.

On the contrary, voxel-based detectors \cite{voxelnet_zhou, second_yan, votr_mao, voxelrcnn_deng} convert point clouds into regular 3D voxels and extract features with convolution operations.
VoxelNet \cite{voxelnet_zhou} first proposed a voxel feature encoding method for point clouds, and SECOND \cite{second_yan} reduced computational cost by introducing efficient sparse convolution \cite{spconv_graham}.
Voxel R-CNN \cite{voxelrcnn_deng} is a typical two-stage detector that takes advantage of voxel representations in the proposal refinement stage via voxel RoI pooling.

To mitigate information loss due to data quantization, voxel-based approaches are often combined with point-level supervision and representations.
Part-A$^{2}$ Net \cite{parta2_shi} and SA-SSD \cite{sassd_he} exhibited remarkable performance using a point-level auxiliary task.
PV-RCNN \cite{pvrcnn_shi} aggregates voxel features at a set of keypoints obtained from the scene using the farthest point sampling (FPS) algorithm, and exploits the keypoint features during the proposal refinement.
Some methods use the points within proposal bounding boxes instead of sampling the points from the entire scene.
For example, CT3D \cite{ct3d_sheng} utilizes a voxel-based backbone, but for refining a proposal, it relies solely on the raw point coordinates within the proposal.
Others \cite{fastpointrcnn_chen, pdv_hu} exploit internal point information alongside the voxel features for RoI feature pooling.
However, the limited number of collected points for distant or occluded objects still poses a challenge to detection performance.

\begin{figure*}
\begin{center}
\includegraphics[width=0.9\linewidth]{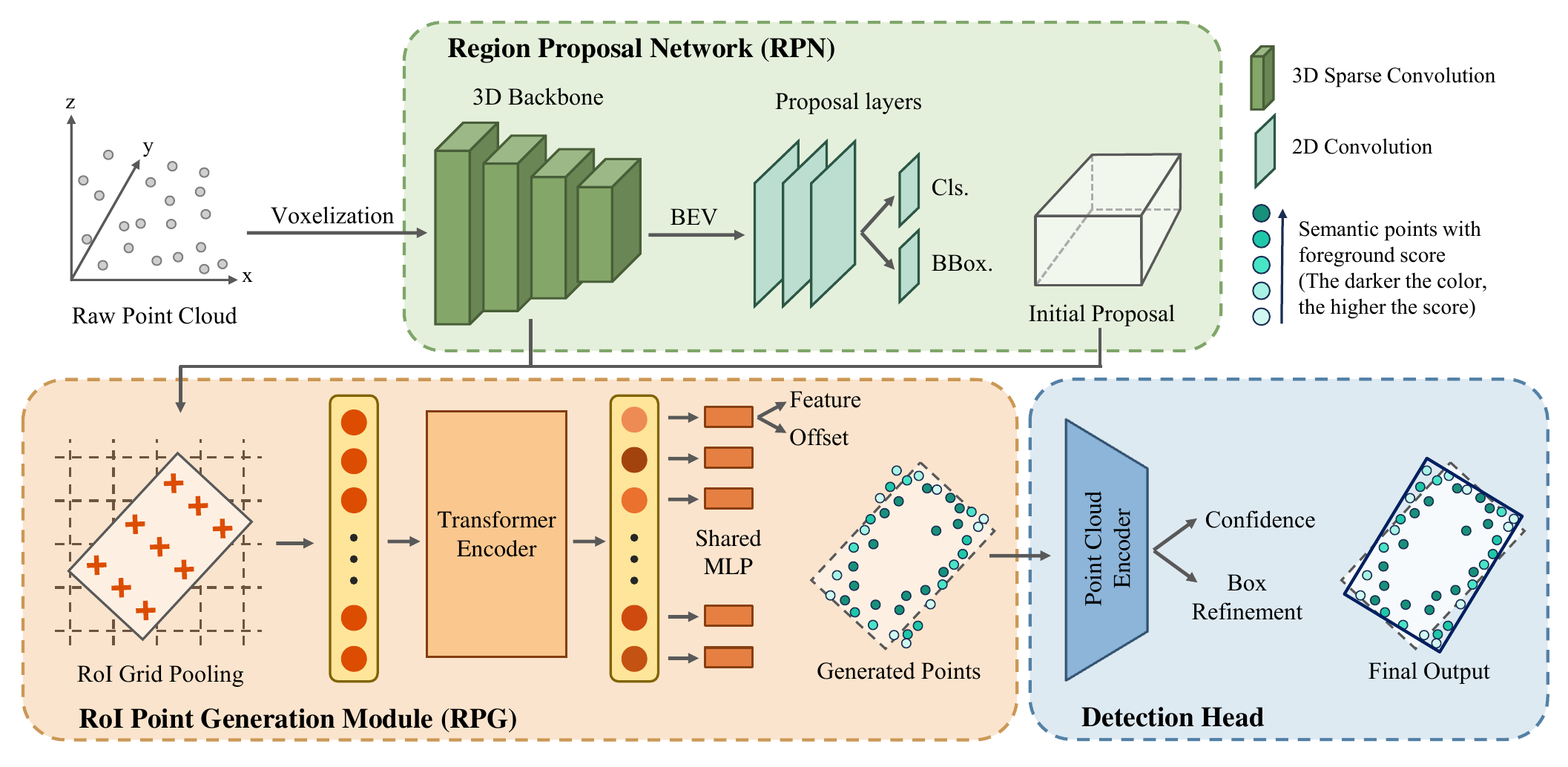}
\end{center}
   \caption{PG-RCNN Overview. The input point cloud is first voxelized and passed through the region proposal network to produce initial bounding box proposals. The RoI point generation module then generates semantic points for each proposal, computing the coordinates offsets and semantic features from the voxel features aggregated at RoI grid points. Finally, the detection head produces the final detection output using the generated semantic point clouds as input.}
\label{fig:2_framework}
\end{figure*}

\subsection{Point Generation for 3D Object Detection}

Several recent works have aimed to overcome the sparsity and incompleteness of LiDAR points by augmenting additional points to the scene.
For example, \cite{mvp_yin, sfd_wu, vpfnet_zhu} incorporate RGB images to generate dense pseudo-LiDAR and virtual points.
In terms of single-modal approaches, PC-RGNN \cite{pcrgnn_zhang} and SIENet \cite{sienet_li} generate dense point clouds of the objects to capture rich spatial information of RoIs.
These methods utilize a pre-trained point cloud completion network that takes raw point coordinates within a initial bounding box proposal as input and outputs a set of point coordinates that form a plausible object shape.
PC-RGNN adds supplementary points to the original points using a GCN-based \cite{gcn_wang} point cloud completion network and encodes their coordinates with a graph neural network.
Meanwhile, SIENet generates dense point clouds with an existing framework called PCN \cite{pcn_yuan}.
Then SIENet extracts features from the generated points using the PointNet++  \cite{pointnet++_qi} encoder and fuses the features with the grid-pooled voxel features to enhance spatial information.
In contrast, our point generation method takes RoI-pooled features as input, capturing the contextual information of the surroundings.
Our approach is also distinguished from SPG \cite{spg_xu}, the unsupervised domain adaptation method that assigns a semantic point to every estimated foreground voxel before feeding into a detector.
SPG does not recover the actual shape of the objects and can be used in parallel with our method as a data pre-processing technique.


\section{PG-RCNN}
PG-RCNN is a two-stage method for 3D object detection composed of a region proposal stage and a proposal refinement stage. 
Figure \ref{fig:2_framework} illustrates the overview of the PG-RCNN framework. 
For the first stage, the region proposal network (RPN) with a voxel-based backbone generates the initial bounding box proposals. 
Our main novelty lies in the second stage, where we introduce the RoI point generation (RPG) module to create a semantic surface point cloud for each proposal. 
The RPG module aggregates the backbone voxel features in a grid and uses a Transformer \cite{transformer_vaswani} encoder to capture the global context of the RoI.
Then an MLP is individually applied to each grid point feature to output the coordinates offset and the semantic feature of the generated point.
Lastly, the detection head produces the final detection output using the bounding box refinement features extracted from the generated point clouds with semantic features.


\subsection{Region Proposal Network}

Following many recent works \cite{pvrcnn_shi, voxelrcnn_deng, ct3d_sheng, pdv_hu}, we adopt SECOND \cite{second_yan} as our RPN for its high efficiency and recall.
The input raw point cloud is first divided into evenly spaced voxels and gradually processed with the 3D backbone network composed of a series of sparse convolution layers, resulting in multiple scales of feature volumes.
The downsampled feature volumes are projected along the Z-axis and converted into a bird's-eye view (BEV) feature map.
The proposal layers use the BEV feature map to produce dense predictions with the classification and box regression branches to generate initial detection output for the later refinement stage.

\subsection{RoI Point Generation Module}

 Previous approaches \cite{sienet_li, pcrgnn_zhang} leverage point-based completion models, using raw points pooled from RoI as input. 
Instead, our RPG module exploits voxel features from the 3D backbone, which contain rich context information about their surroundings.

We begin by dividing a region proposal into $G \times G \times G$ regular sub-voxels, using the center coordinates of these sub-voxels as grid points.
Then, we use a method from Voxel R-CNN \cite{voxelrcnn_deng} to aggregate voxel features at the grid points.
Specifically, a grid point $\mathbf{g}_i$ is quantified into a voxel, so that the neighboring voxels are efficiently obtained by indices translation.
Using a PointNet++ \cite{pointnet++_qi} module, we aggregate its feature $\mathbf{f}_{\mathbf{g}_i}$ from the sampled set of neighboring voxels $\Gamma_i = \{\mathbf{v}_i^1, \mathbf{v}_i^2, \cdots, \mathbf{v}_i^K\}$ as follows:
\begin{equation}
\mathbf{f}_{\mathbf{g}_{i}} = {MaxPool} \left( \{{\mathcal{A}^{agg}([\mathbf{v}_i^k - \mathbf{g}_i; \mathbf{f}_{\mathbf{v}_i^k}])}\}_{k=1}^K \right),
\end{equation}
where $\mathcal{A}^{agg}(\cdot)$ represents the MLP for feature aggregation, $\mathbf{v}_i^k - \mathbf{g}_i$ represents relative coordinates, and $\mathbf{f}_{\mathbf{v}_i^k}$ is the feature of voxel $\mathbf{v}_i^k$. The RPG module aggregates voxel features from feature volumes of the last three stages in the 3D backbone network and concatenates the multi-scale features.

The feature pooled at each grid point contains local information about its surroundings but lacks RoI-level context information for estimating object shapes.   
To capture the long-range dependencies between the grid points, we further process the features with a Transformer encoder.
In Section \ref{sec:4.4}, we demonstrate the effectiveness of utilizing the Transformer encoder in enhancing object detection performance.
The refined grid point feature $\Tilde{\mathbf{f}}_{\mathbf{g}_i}$ is formulated as
\begin{equation}
    \Tilde{\mathbf{f}}_{\mathbf{g}_i} = \mathcal{T}(\mathbf{f}_{\mathbf{g}_i}, \mathbf{\delta}_{\mathbf{g}_i}),
\end{equation}
where $\mathbf{\delta}_{\mathbf{g}_i}$ is the positional encoding, and $\mathcal{T}(\cdot)$ is a standard Transformer encoder. To encode positional information, we apply a shallow feedforward neural network (FFN) to the relative coordinates of the grid point with respect to the region proposal bounding box, as described in \cite{ct3d_sheng}:
\begin{equation}
    \mathbf{\delta}_{\mathbf{g}_i} = \mathcal{A}^{pos}([\mathbf{g}_i - \mathbf{r}^c; \mathbf{g}_i - \mathbf{r}^1; \mathbf{g}_i - \mathbf{r}^2; \cdots; \mathbf{g}_i - \mathbf{r}^8]),
\end{equation}
where $\mathcal{A}^{pos}(\cdot)$ represents the FFN, $\mathbf{r}^c$ is the center, and $\mathbf{r}^{1,2,\cdots,8}$ are the eight corners of the bounding box.

Finally, a two-layer MLP $\mathcal{A}^{gen}(\cdot)$ is applied to the refined features to generate the offset $\mathbf{o}_i$ from the grid point, as well as the semantic feature of the generated point $\mathbf{f}^{se}_{\mathbf{p}_i}$:
\begin{equation}
    [\mathbf{o}_i; \mathbf{f}^{se}_{\mathbf{p}_i}] = \mathcal{A}^{gen}(\Tilde{\mathbf{f}}_{\mathbf{g}_i}).
\end{equation}
The generated point's coordinates $\mathbf{p}_i = (x_i,y_i,z_i)$ can be calculated as  $\mathbf{g}_i + \mathbf{o}_i$, and the foreground score $s_i$ for each generated point is calculated by applying a linear projection $A$ and a sigmoid function $\sigma$ to its semantic feature, \ie, 
\begin{equation}
    s_i = \sigma(A\mathbf{f}^{se}_{\mathbf{p}_i}).
\end{equation}

\subsection{Detection Head}

Our detection head is inspired by the design of PointRCNN \cite{pointrcnn_shi} where it uses the PointNet++ encoder to extract refinement features from semantic point clouds.
For every generated point, we obtain the local spatial feature $\mathbf{f}^{sp}_{\mathbf{p}_i}$ with an MLP $\mathcal{A}^{loc}$ as follows:
\begin{equation}
    \mathbf{f}^{sp}_{\mathbf{p}_i} = \mathcal{A}^{loc} \left( \left[x_i^c, y_i^c, z_i^c, d_i, s_i \right]\right).
\end{equation}
Here, $(x_i^c, y_i^c, z_i^c)$ are the coordinates of the generated point $\mathbf{p}_i$ in the canonical coordinates system of the bounding box, and $d_i=\sqrt{x_i^2 + y_i^2 + z_i^2}$ is the depth of the point. 
The canonical transformation facilitates robust local spatial feature learning.
However, the transformation causes the inevitable loss of points' depth information, so we append $d_i$ as an additional feature. 
The estimated foreground score $s_i$ is also appended as the feature that represents the significance of the generated point.
We merge $\mathbf{f}^{sp}_{\mathbf{p}_i}$ and $\mathbf{f}^{se}_{\mathbf{p}_i}$ for each point $\mathbf{p}_i$, and feed the point set with features into the PointNet++ encoder to obtain the refinement feature for RoI $\mathbf{f}^{r}$:
\begin{equation}
   \mathbf{f}^{r} = \mathcal{P}\left( \{\mathbf{p}_i\}_{i=1}^{G^3}, \{[\mathbf{f}^{sp}_{\mathbf{p}_i};\mathbf{f}^{se}_{\mathbf{p}_i}]\}_{i=1}^{G^3}\right), 
\end{equation}
where $\mathcal{P}(\cdot)$ denotes the PointNet++ encoder taking the set of point coordinates and the corresponding feature set as input.
The RoI feature serves as the input for confidence classification and bounding box refinement branches, resulting in the final detection output.

\subsection{Training Losses}

PG-RCNN is an end-to-end model trained with the summation of the region proposal loss $\mathcal{L}_{\mathrm{RPN}}$, the proposal refinement loss $\mathcal{L}_{\mathrm{head}}$, and the point generation loss $\mathcal{L}_{\mathrm{RPG}}$:
\begin{equation}
    \mathcal{L}_{total} = \mathcal{L}_{\mathrm{RPN}} + \mathcal{L}_{\mathrm{head}} + \mathcal{L}_{\mathrm{RPG}}. 
\end{equation}

$\mathcal{L}_{\mathrm{RPN}}$ and $\mathcal{L}_{\mathrm{head}}$ are conventional training losses for two-stage detectors calculated with the outputs of the RPN and the detection head, respectively.
Both losses are composed of a classification and a regression term.
The classification targets are assigned based on the intersection-over-union (IoU) of the anchors and the proposals with the ground truth bounding boxes. Only foreground anchors and proposals contribute to the regression losses, using the regression target given by their ground truth residuals.
Focal Loss \cite{focal_yun} is used for the RPN's classification branch output, while binary cross-entropy loss is used for the detection head's confidence branch output. For the regression loss, we use the smooth-L1 loss for both losses.

$\mathcal{L}_{\mathrm{RPG}}$ is an auxiliary loss term that specifically supervises point generation, calculated with the RPG module outputs:
\begin{equation}
    \mathcal{L}_{\mathrm{RPG}} = \mathcal{L}_{score} + \mathcal{L}_{offset}.
\end{equation}
$\mathcal{L}_{score}$ is a point-level segmentation loss that governs the foreground scores of generated points.
We assign segmentation labels to generated points by checking if they are inside a ground-truth bounding box.
Since we generate $G^3$ points for each proposal, calculating loss at all generated points would be computationally expensive.
We select $N_p$ points from the scene using the FPS algorithm and apply Focal Loss on the sampled points, \textit{i.e.},
\begin{equation}
    \mathcal{L}_{score} = -\frac{1}{N_p} \sum_{j}{(1-s_j)^\gamma \log{s_j}}
\end{equation}
where $s_j, j=1, 2, \cdots N_p$ are the foreground score of the sampled points.

\begin{figure}[t]
\begin{center}
   \includegraphics[width=\linewidth]{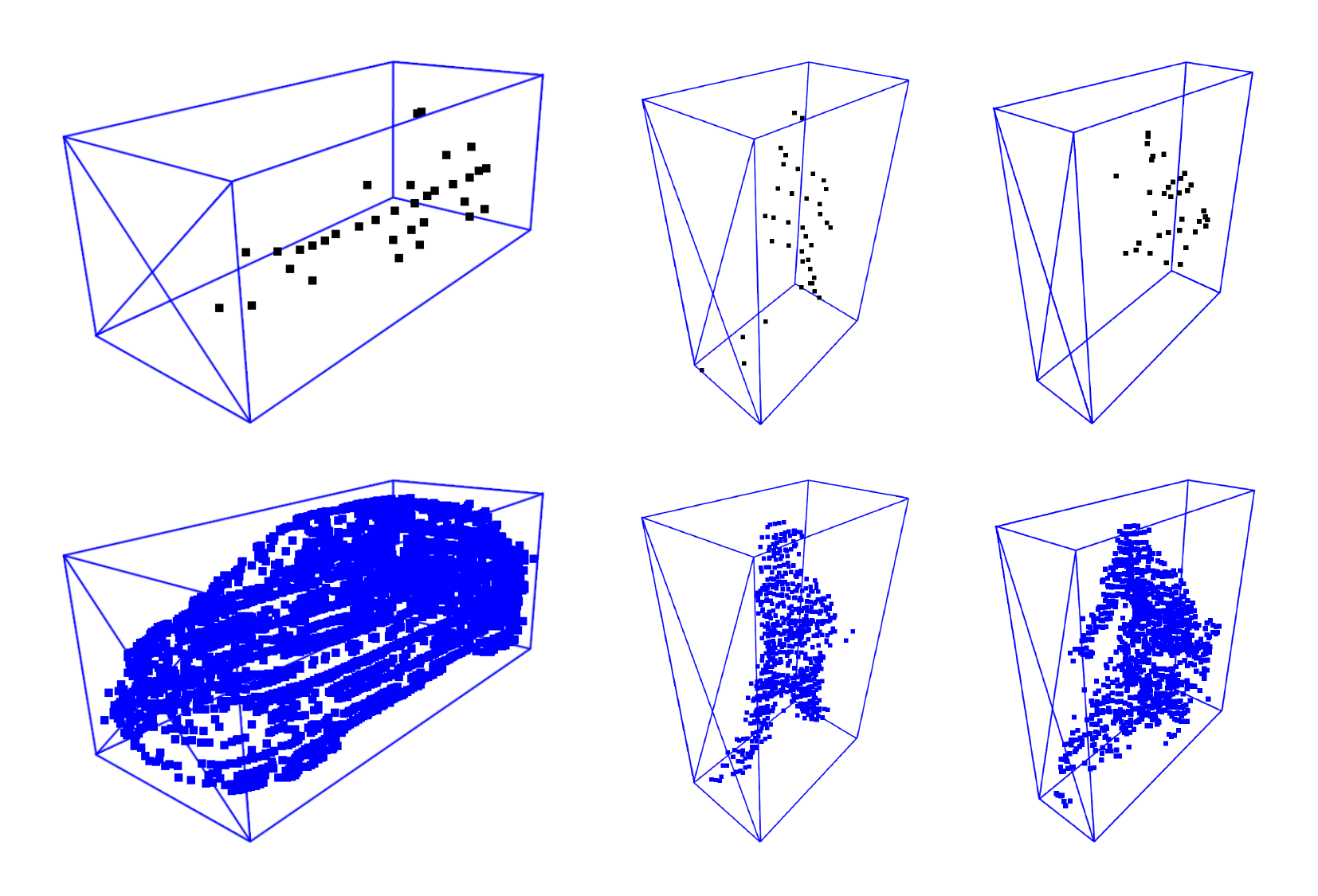}
\end{center}
   \caption{Examples of completed point clouds for a car, pedestrian, and cyclist. Original point clouds (top) and their corresponding completed point clouds for supervising point generation (bottom).}
\label{fig:3_target}
\end{figure}

On the other hand, $\mathcal{L}_{offset}$ supervises the shape of the generated point clouds. 
Since the KITTI dataset does not provide complete point clouds of the object instances, previous approaches \cite{pcrgnn_zhang, sienet_li} used external datasets such as ShapeNet \cite{shapenet} to train their point cloud completion network in advance. 
Instead, we exploit other object instances within the provided dataset to approximate the complete shape of the object.
Specifically, we use the approximation method proposed in \cite{btc_xu}.
We first search for other objects of the same class that have similar bounding boxes and point distributions.
Then we combine the point sets of two best-matching objects with the original points and produce a dense point cloud.
For cars and cyclists, we assume symmetry along the object's heading axis and mirror the points accordingly.
Figure \ref{fig:3_target} displays an example of a completed point cloud for each class.
Using the completed point clouds as generation targets, we employ Chamfer Distance on foreground proposals as follows:
\begin{multline}
    \mathcal{L}_{offset} = \frac{1}{N_{fp}}\sum_{r}\left(\dfrac{1}{|\mathbf{P}_r|} \sum_{\mathbf{x} \in \mathbf{P}_r}\min_{\mathbf{y}\in \mathbf{P}_r^*}||\mathbf{x}-\mathbf{y}||^2_2 + \right. \\
    \left. \dfrac{1}{|\mathbf{P}_r^*|} \sum_{\mathbf{y} \in \mathbf{P}_r^*}\min_{\mathbf{x}\in \mathbf{P}_r}||\mathbf{y}-\mathbf{x}||^2_2 \right),  
\end{multline}
where $N_{fp}$ is the number of foreground proposals, and $\mathbf{P}_r$ and $\mathbf{P}_r^*$ are the generated and the target point cloud of the $r$-th foreground proposal where $r=1, 2, \cdots, N_{fp}$, respectively.

\section{Experiments}

In this section, we conduct a comprehensive analysis on the KITTI dataset \cite{kitti} to verify the effectiveness of PG-RCNN and its components.
In Sec\onedot \ref{sec:4.2}, we evaluate PG-RCNN on the competitive benchmark and compare the performance with the state-of-the-art methods.
Furthermore, we qualitatively compare our point generation results with a prior point cloud completion approach \cite{sienet_li} in Sec\onedot \ref{sec:4.3}. Extensive ablation studies in Sec\onedot \ref{sec:4.4} validate our design. 
We also conducted experiments on the Waymo Open Dataset \cite{waymo}, another popular autonomous driving dataset.
Please refer to the supplementary materials for  experiments on Waymo Open Dataset. 

\subsection{Experimental Setup}\label{sec:4.1}
\paragraph{KITTI Dataset.}
The KITTI dataset provides 7,481 annotated training samples and 7,518 testing samples.
Following \cite{kitti_split_chen}, we split the original training data into 3,712 and 3,769 samples for training and validation, respectively.
We detect three object classes: cars, cyclists, and pedestrians.



\paragraph{Network Architecture.}
We limit the detection range as [0m, 70.4m] for the X-axis, [-40m, 40m] for the Y-axis, and [-3m, 1m] for the Z-axis.
To process this data, the raw point clouds are divided into voxels of size (0.05m, 0.05m, 0.1m) along each axis.
The feature dimensions of the 3D backbone are (16, 32, 48, 64) across four stages, while the proposal layer's feature dimensions are (64, 128).
In the RoI grid pooling step, the dimension of each grid's feature $\mathbf{f}_{\mathbf{g}_{i}}$ is set to 96 with a grid size $G$ of 6.
We use a single-layer Transformer encoder with a hidden feature dimension of 384.
The semantic feature vector $\mathbf{f}^{se}_{\mathbf{p}_{i}}$ and local spatial feature vector $\mathbf{f}^{sp}_{\mathbf{p}_{i}}$ of generated points have 32 and 64 dimensions, respectively.
For each proposal, the point cloud encoder in the detection head extracts an RoI feature vector $\mathbf{f}^{r}$ of dimension 256.
Overall, PG-RCNN use lighter MLP layers than the motivational works \cite{second_yan, voxelrcnn_deng, ct3d_sheng, pointrcnn_shi}, allowing our model to be significantly more efficient than the previous methods (please refer to Table \ref{table:kitti_val}).

\begin{table*}[ht!]
 \caption{Comparison with state-of-the-art methods on the KITTI \textit{val} set. $^\dag$ denotes the re-implemented model, in which we replaced the first stage detector with RPN of ours. Latency is reported with the average inference time on a single NVIDIA RTX 3090 GPU. The best performance value is in \textbf{bold}, second-best is \underline{underlined}.}
  \centering
  \label{table:kitti_val}
  \vspace{0.1cm}
  \begin{tabular}{l||c|c|c c c|c c c|c c c}
    \hline
    \multirow{2}{*}{Method} &
    Param.&
    Latency&
    \multicolumn{3}{c|}{Car 3D AP$_{R40}$} &
    \multicolumn{3}{c|}{Ped. 3D AP$_{R40}$} &
    \multicolumn{3}{c}{Cyc. 3D AP$_{R40}$} \\
    &(M) & (ms)& Easy & Mod. & Hard & Easy & Mod. & Hard & Easy & Mod. & Hard  \\
    \hline
    SECOND \cite{second_yan}&5.33&\underline{59.9}& 90.55 & 81.61 & 78.56 & 55.94 & 51.15 & 46.17 & 82.97 & 66.74 & 62.78 \\ 
    PointPillars \cite{pointpillars_lang}&4.83&\textbf{36.5}& 87.75 & 78.41 & 75.19 &  57.30 & 51.42 & 46.87 & 81.57 & 62.93 & 58.98 \\ 
    PointRCNN \cite{pointrcnn_shi}&\underline{4.04}&171.5& 91.39 & 80.53 & 78.05 & 62.41 & 55.70 & 49.01 & \underline{92.56} & 73.13 & 68.81\\ 
    PV-RCNN \cite{pvrcnn_shi}&13.12& 103.5& 92.10 &84.36& 82.48&\underline{64.26}&56.67&51.91&88.88&71.95&66.78 \\ 
    CT3D \cite{ct3d_sheng} &7.85&142.3& 92.34&84.97&\underline{82.91}&	61.05&55.57&51.10&89.01&71.88&67.91 \\
    PC-RGNN \cite{pcrgnn_zhang} & 21.43 & 152.4 & 90.94&81.43&80.45 &$-$&$-$&$-$&$-$&$-$&$-$  \\ 
    Voxel-RCNN \cite{voxelrcnn_deng}&7.59& 93.2&91.72 & 83.19 & 78.60 &$-$&$-$&$-$&$-$&$-$&$-$ \\ 
    PDV \cite{pdv_hu}&12.86&161.5& 92.44 & 85.05 & 82.77 & 63.89 & \underline{57.41} & \underline{52.56} & 91.78 & \textbf{75.95} & \textbf{71.36} \\
    SIENet \cite{sienet_li} & 24.62 & 120.8 & \underline{92.49} & \textbf{85.43} & \textbf{83.05} & $-$& $-$& $-$&$-$&$-$&$-$ \\
    SIENet$^\dag$ &21.03&117.6& 91.96 & 84.45 & 82.64& $-$& $-$& $-$&$-$&$-$&$-$ \\
    \hline
    PG-RCNN (Ours) & \textbf{2.28} & 60.1 & \textbf{92.73}&\underline{85.26}&82.83&\textbf{68.44}&\textbf{60.63}&\textbf{55.36}&\textbf{93.84}&\underline{74.85}&\underline{70.15}\\
    \hline
\end{tabular}
\end{table*}
\begin{table*}[ht!]
  \centering
  \caption{Performance comparison on the KITTI \textit{test} set with AP under 40 recall positions. Most of the results were obtained from the KITTI test server, while $^\ddag$ indicates the model whose result was obtained from the paper because it was not reported to the server. The best performance value is in \textbf{bold}, second-best is \underline{underlined}.}
  \label{table:kitti_test}
  \vspace{0.1cm}
  \resizebox{\textwidth}{!}{%
  \begin{tabular}{l||c c c|c c c|c c c|c c c}
    \hline
    \multirow{2}{*}{Method} &
    \multicolumn{3}{c|}{Car 3D AP$_{R40}$} &
    \multicolumn{3}{c|}{Car BEV AP$_{R40}$} &
    \multicolumn{3}{c|}{Cyc. 3D AP$_{R40}$} &
    \multicolumn{3}{c}{Cyc. BEV AP$_{R40}$} \\
    & Easy & Mod. & Hard & Easy & Mod. & Hard & Easy & Mod. & Hard & Easy & Mod. & Hard  \\
    \hline
    SECOND$^\ddag$ \cite{second_yan} & 83.13 & 73.66 & 66.20 & 88.07 & 79.37 & 77.95 & 70.51 & 53.85 & 46.90 & 73.67 & 56.04 & 48.78 \\
    PointPillars \cite{pointpillars_lang} & 82.58 & 74.31 & 68.99 & 90.07 & 86.56 & 82.81 & 77.10 & 58.65 & 51.92 & 79.90 & 62.73 & 55.58 \\
    PointRCNN \cite{pointrcnn_shi} & 86.96 & 75.64 & 70.70 & 92.13 & 87.39 & 82.72 & 74.96 & 58.82 & 52.53 & 82.56 & 67.24 & 60.28 \\
    PartA$^{2}$ \cite{parta2_shi} & 87.81 & 78.49 & 73.51 & 91.70 & 87.79 & 84.61 & 79.17 & 63.52 & 56.93 & 83.43 & 68.73 & 61.85 \\
    Point-GNN \cite{pointgnn_shi} & 88.33 & 79.47 & 72.29 & 93.11 & 89.17 & 83.90 & 78.60 & 63.48 & 57.08 & 81.17 & 67.28 & 59.67 \\
    3DSSD \cite{3dssd_yang} & 88.36 & 79.57 & 74.55 & 92.66 & 89.02 & 85.86 & 82.48 & 64.10 & 56.90 & \underline{85.04} & 67.62 & 61.14 \\
    
    PV-RCNN \cite{pvrcnn_shi} & 90.25 & 81.43 & 76.82 & \textbf{94.98} & \textbf{90.65} & 86.14 & 78.60 & 63.71 & 57.65 & 82.49 & 68.89 & 62.41 \\
    CT3D \cite{ct3d_sheng} & 87.83 & 81.77 & 77.16 & 92.36 & 88.83 & 84.07 & $-$ & $-$ & $-$ & $-$ & $-$ & $-$ \\
    PC-RGNN$^\ddag$ \cite{pcrgnn_zhang} & 89.13 & 79.90 & 75.54 & \underline{94.91} & 89.62 & \textbf{86.57} & $-$ & $-$ & $-$ & $-$ & $-$ & $-$ \\
    Voxel R-CNN \cite{voxelrcnn_deng} & \textbf{90.90} & 81.62 & 77.06 & 94.85 & 88.83 & 86.13 & $-$ & $-$ & $-$ & $-$ & $-$ & $-$ \\
    BtcDet \cite{btc_xu} & \underline{90.64} & \textbf{82.86} & \textbf{78.09} & 92.81 & 89.34 & 84.55 & \underline{82.81} & \textbf{68.68} & \textbf{61.81} & 84.48 & \textbf{71.76} & \textbf{64.70} \\
    PDV \cite{pdv_hu} & 90.43 & 81.86 & \underline{77.36} & 94.56 & \underline{90.48} & 86.23 & \textbf{83.04} & 67.81 & 60.46 & \textbf{85.54} & \underline{71.31} & 64.40 \\
    SIENet \cite{sienet_li} & 88.22 & 81.71 & 77.22 & 92.38 & 88.65 & 86.03 & $-$ & $-$ & $-$ & $-$ & $-$ & $-$ \\
    \hline
    PG-RCNN (Ours) & 89.38&\underline{82.13}&77.33&93.39&89.46&\underline{86.54}&82.77&\underline{67.82}&\underline{61.25}&84.94&70.65&64.03 \\
    \hline
  \end{tabular}
  }
\end{table*}
\paragraph{Training Details.}
For data augmentation, we apply widely employed strategies, including random flipping along the X-axis, global scaling, global rotation around the Z-axis, and ground truth sampling.
Please refer to OpenPCDet \cite{openpcdet} for detailed configurations since we used the toolbox for all our experiments.
PG-RCNN is trained using the Adam optimizer \cite{adam_kingma} with a one-cycle policy for 80 epochs with an initial learning rate of 0.01.
We used 4 NVIDIA RTX 3090 GPUs to train our network with a batch size of 16, and the training time was less than 5 hours.
$N_p$, the number of points used to calculate $\mathcal{L}_{score}$, is set to 2,048.

\subsection{Comparison with State-of-the-Arts}\label{sec:4.2}

We trained our model on the \textit{train} set and tuned the hyperparameters based on the \textit{val} set evaluation results.
To submit the detection results on KITTI official test server, we trained the model using all annotated \textit{train}+\textit{val} samples.
All results are evaluated by the mean average precision (AP) calculated with 40 recall positions (R40), using IoU thresholds of 0.7 for cars, and 0.5 for pedestrians and cyclists.
The evaluation results are reported on three levels of difficulties: easy, moderate, and hard.

\begin{figure*}[t]
\begin{center}
   \includegraphics[width=\textwidth]{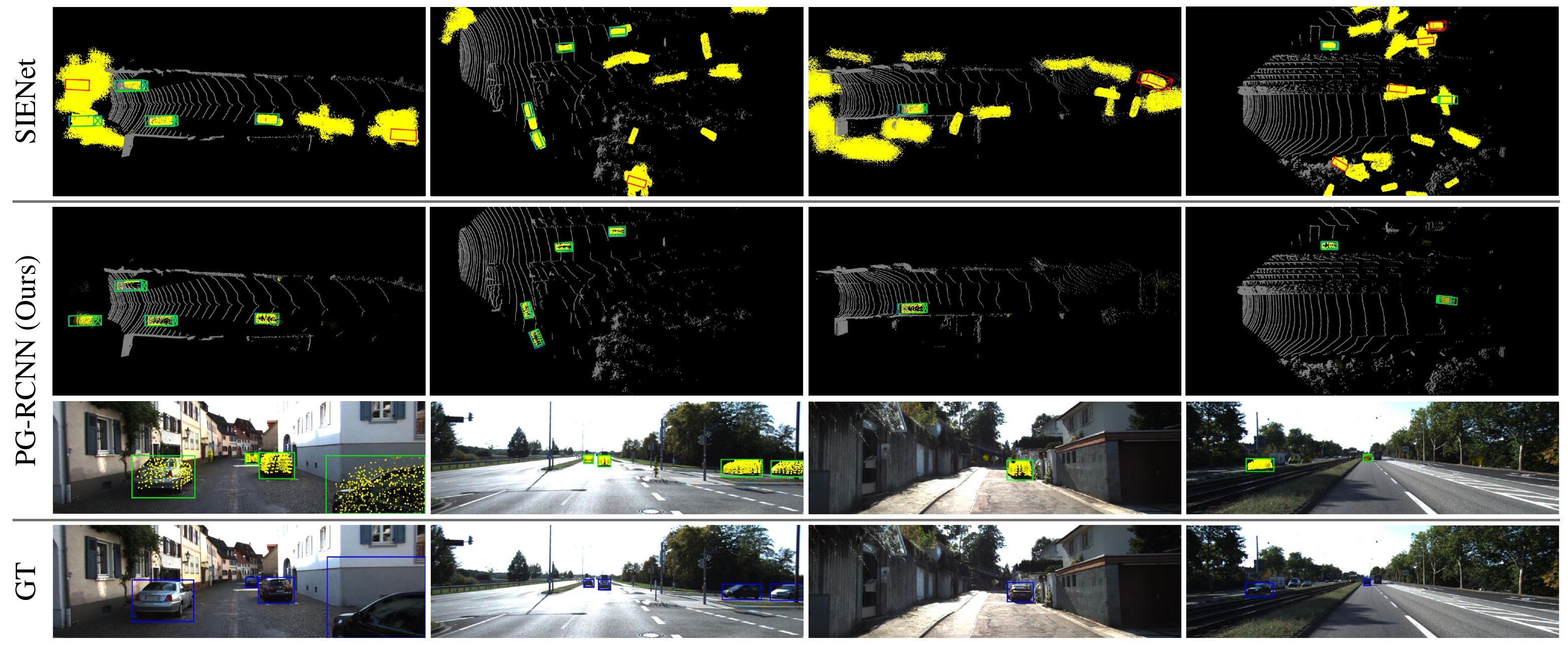}
\end{center}
   \caption{The point generation and detection results of SIENet and PG-RCNN (ours) on KITTI \textit{val} samples. The generated points, true positive predictions, false positive predictions, and ground truth bounding boxes are highlighted in yellow, green, red, and blue, respectively. }
\label{fig:4}
\end{figure*}

Table \ref{table:kitti_val} summarizes the performance comparison of PG-RCNN on KITTI \textit{val} set with the state-of-the-art models that officially released the trained weights.
PG-RCNN shows the best or second-best performance for all classes and difficulties, except for the car class on a hard difficulty, where we achieved the third-best performance.
The previous point cloud completion approach, SIENet \cite{sienet_li} surpasses our model on car class for moderate and hard difficulties.
We believe this is due to its advanced region proposal network, and we re-implemented SIENet with general RPN from SECOND \cite{second_yan} as our model to fairly compare the effect of point generation on the refinement stage.
In this case, it can be observed that PG-RCNN outperforms the model (denoted as SIENet$^\dag$ in Table \ref{table:kitti_val}) at all levels of the car class, implying that our refinement method is more effective.
Moreover, PG-RCNN exhibits remarkably superior efficiency compared to recent methods.
Notably, our model has over 9 times fewer parameters in to previous point cloud completion approaches.
PG-RCNN also has a low inference time demand, comparable to a single-stage detector \cite{second_yan}.

PG-RCNN also obtains competitive detection performance on the KITTI \textit{test} set, as summarized in Table \ref{table:kitti_test}.
We ranked second or third place on the 3D detection results except on the easy level for the car class.
In comparison to previous point cloud completion approaches \cite{pcrgnn_zhang, sienet_li}, PG-RCNN consistently outperforms the competitors on 3D detection performances for cars on all levels.
Although our method's performance falls short on certain metrics when compared to the most recent publications, PG-RCNN still exhibits a remarkable trade-off in terms of high efficiency. 
However, we hypothesize that our model's lack of scalability in \textit{test} set evaluation results from its lightweight nature. 
In our future work, we plan to explore the use of a more sophisticated detection head to improve detection performance on larger datasets.

\subsection{Analysis on Point Generation Results}\label{sec:4.3}
Here, we compare the qualitative results of the proposed method on KITTI \textit{val} data with a previous point cloud completion method, SIENet \cite{sienet_li}.
To fully focus on the effect of point generation in the refinement stage, we compare our model with SIENet$^\dag$ we presented in Table \ref{table:kitti_val}.

\begin{figure}[t]
\begin{center}
   \includegraphics[width=\linewidth]{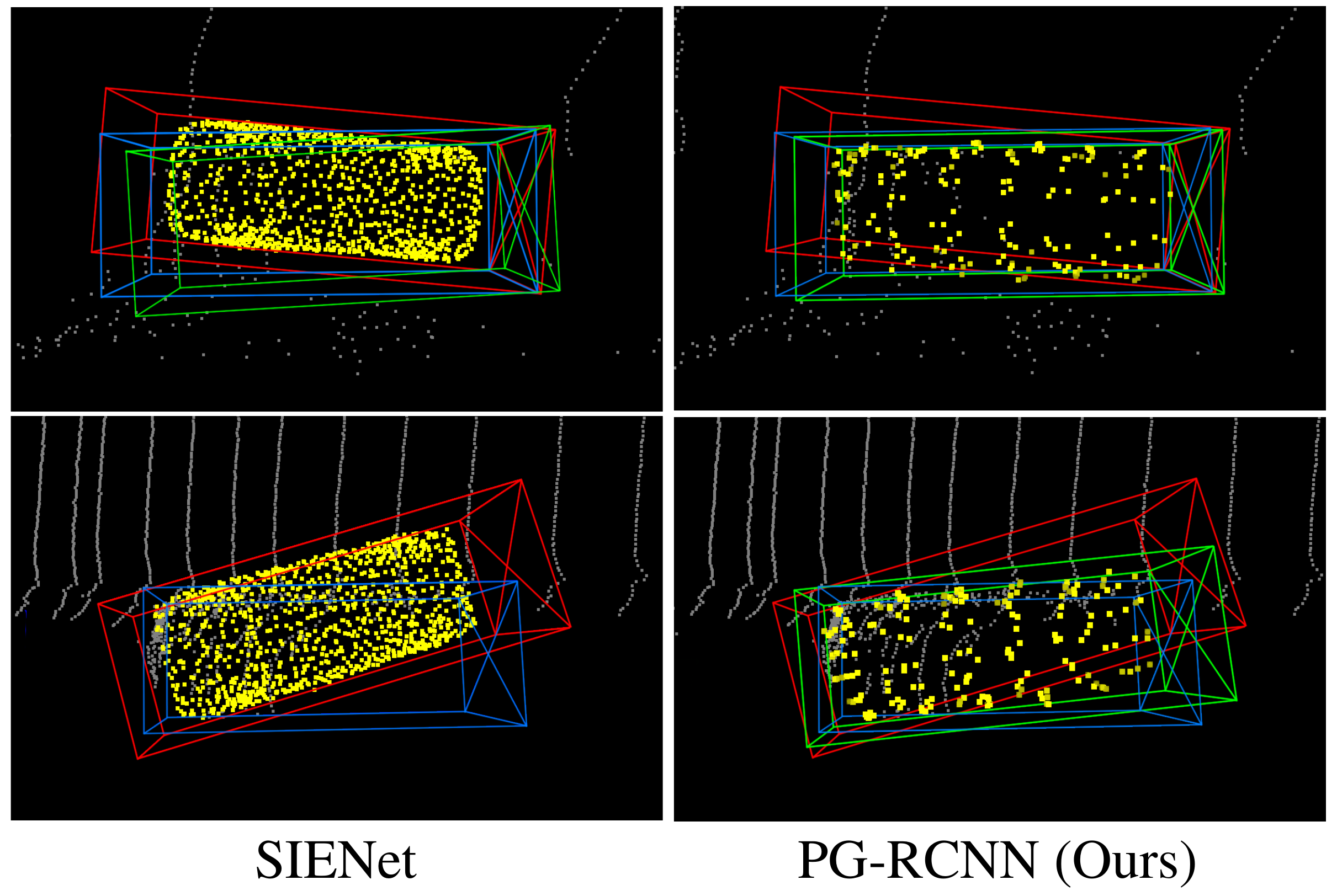}
\end{center}
   \caption{The point generation and refinement results for a misaligned proposal. The generated points, initial proposal, refinement results, and ground truth are highlighted in yellow, red, green, and blue, respectively. }
\label{fig:5}
\end{figure}
Figure \ref{fig:4} illustrates some of the point generation and detection results of SIENet and PG-RCNN.
The foreground score of each point is expressed with its opacity in the figure.
Since the point cloud completion network of SIENet only produces spatial coordinates, we set the foreground score of all its generated points to 1.
The top two rows of Fig\onedot \ref{fig:4} display the outputs in a bird's-eye-view.
Observations show that SIENet indiscriminately generates point clouds for all region proposals, and results more false positive predictions than ours.
In contrast, our method presents high-confidence foreground points only at true positive bounding boxes.
This suggests that considering foreground probabilities of generated points can effectively regularize producing false positive detection results.
The third row of Fig\onedot \ref{fig:4} exhibits the projections of the outputs of our model onto the images, showing that the generated points are well-aligned with the foreground objects.
The results suggest that PG-RCNN can not only detect objects but also successfully estimate their actual shape.

To further investigate the effectiveness of our point generation method, we compare how the point cloud completion network of SIENet and our RPG module behave in the same situation.
We artificially composed misaligned proposals by slightly distorting ground truth bounding boxes, and provided them to both models.
Figure \ref{fig:5} illustrates some of the refinement stage outputs of SIENet and ours.
Although SIENet creates a denser point cloud than ours, it does not align with existing foreground points nor get outside of the proposal bounding box.
In the top example of Fig\onedot \ref{fig:5}, SIENet refined the proposal towards the ground truth bounding box.
However, this prediction did not align with the point generation result.
Moreover, in the bottom example, SIENet was unable to make a confident final prediction with the generated point cloud.
This indicates the generated points are \textit{pointless} for proposal refinement.
On the contrary, points generated with our method actively move outside the initial proposal, attempting to capture the actual foreground object surface.
The generated points mostly fit within the ground truth bounding box, and the final detection bounding boxes are predicted accordingly.
The intuitive comparisons show that our point generation results better serve a purpose for addressing misaligned proposals.

\subsection{Ablation Studies}
\label{sec:4.4}
To verify the effectiveness of the proposed method, we conduct extensive ablation studies on the KITTI \textit{val} set.
\vspace{-2ex}
\paragraph{Point Generation Loss.} 
Our RPG module is trained with the supervision of two losses, $\mathcal{L}_{score}$ and $\mathcal{L}_{offset}$, which assign semantic and geometric attributes to the generated points, respectively.
To investigate the impact of these supervisions on object detection performance, we ablated each loss term and compared the 3D detection performances on car objects.
In the absence of $\mathcal{L}_{score}$, a uniform foreground score of $s_i = 1$ is allocated for all generated points.
Similarly, a fixed offset $\mathbf{o}_i = (0,0,0)$ is used for the absence of $\mathcal{L}_{offset}$, indicating that all generated points were located at the grid centers.
Table \ref{table:ablation1} summarizes the results of the experiments.
Our model showed a consistent performance drop when we did not employ $\mathcal{L}_{score}$, resulting in a decline of 0.82\% in mAP.
This reveals that assigning semantic information to generated points is an important feature of our method.
Similarly, when $L_{offset}$ was not utilized, the mAP decreased by 0.61\%. This result demonstrates the necessity of spatial supervision, which provides a beneficial trait of shape-awareness for refinement.

\vspace{-2ex}
\paragraph{RoI Point Generation Module.}
In this ablation study, we justify the decision choices regarding the components of the RPG module. Here we compare the 3D detection performances of three classes on the moderate level.
First, we examined the performance gain brought by the Transformer \cite{transformer_vaswani} encoder.
Second, we verified the method for deriving generate points' coordinates, comparing the case of using the RoI center and grid points as the reference point for predicting offsets.
Table \ref{table:ablation2} summarizes the results of the experiment.
The use of the Transformer encoder resulted in a gain of over 1\% in AP for all classes, highlighting the advantage of accessing RoI-level contextual information by employing this component.
The results demonstrate that using grid points as the offset center can significantly improve detection performance for all classes, as compared to using the RoI center as the offset center.

\begin{table}[t!]
  \caption{Performance comparison of adopting different point generation loss.}
  \centering
  \label{table:ablation1}
  \vspace{0.1cm}
  \begin{tabular}{c c||c c c|c}
    \hline
    \multirow{2}{*}{$\mathcal{L}_{score}$}&
    \multirow{2}{*}{$\mathcal{L}_{offset}$}&
    \multicolumn{3}{c|}{Car 3D AP$_{R40}$}&
    \multirow{2}{*}{mAP}\\
    \cline{3-5}
    & & Easy & Mod. & Hard &\\
    \hline
    &\checkmark& 92.31 & 83.97 & 82.03 & 86.10 \\
    \checkmark&& 91.90 & 84.72 & 82.33 &86.31 \\
    \checkmark&\checkmark& \textbf{92.81}& \textbf{85.22} & \textbf{82.74} & \textbf{86.92}\\
    \hline
\end{tabular}
\end{table}
\begin{table}[t!]
  \caption{Performances comparison of different implementations of RoI point generation module.}
  \centering
  \label{table:ablation2}
  \vspace{0.1cm}
  \begin{tabular}{c c||c c c}
    \hline
    \multirow{2}{*}{$\mathcal{T}(\cdot)$}&
    \multirow{2}{*}{offset center}&
    \multicolumn{3}{c}{3D AP$_{R40}$ (Mod.)}\\
    \cline{3-5}
    & & Car & Ped. & Cyc.\\
    \hline
    &grid points& 82.37 & 58.39 & 73.79\\ 
    \checkmark&RoI center& 81.94 & 56.44 & 72.80\\
    \checkmark&grid points& \textbf{85.22}& \textbf{60.45}& \textbf{74.84}\\
    \hline
\end{tabular}
\end{table}

\vspace{-1ex}
\section{Conclusion}
\vspace{-1ex}
In this paper, we present a novel two-stage detector called Point Generation R-CNN (PG-RCNN), that address the LiDAR-based 3D object detection problem by generating semantic surface points of the foreground objects.
PG-RCNN is distinguished from existing point cloud completion approaches in three aspects.
First, our RoI point generation (RPG) module takes grid-pooled backbone features instead of raw coordinates of the points in RoI.
Therefore, it can process the contextual information of the proposal's surrounding and estimate the actual shape and displacement of foreground objects.
Secondly, our method discriminates the generated points by giving them semantic features that represent foreground probabilities, allowing the model to distinguish incorrect proposals during the refinement stage.
Lastly, the RPG module is jointly trained with the rest of the PG-RCNN components without demanding an external dataset for supervision.
Consequently, the proposed method provides intuitive and informative point clouds with semantic features for accurate object detection.
PG-RCNN achieves highly competitive performance on the KITTI dataset while exhibiting significantly better efficiency than previous methods.




\vspace{-1ex}
\section*{Acknowledgements}
This research was financially supported by the Institute of Civil Military Technology Cooperation funded by the Defense Acquisition Program Administration and Ministry of Trade, Industry and Energy of Korean government under grant No. 22-SN-AU-09

{\small
\bibliographystyle{ieee_fullname.bst}
\bibliography{egbib.bib}
}

\end{document}